\documentclass[11pt]{article}
\usepackage[final]{nlp4call}
\usepackage{times}
\usepackage{xurl}
\usepackage{latexsym}
\usepackage{amssymb}
\usepackage{booktabs}
\usepackage{graphicx}

\title{Acoustic and Lexical Signals for Native Language Identification of Language Learners}

\author{
  \textbf{Ahmet Yavuz Uluslu}$^{1,2}$ \quad
  \textbf{Mengjie Qian}$^1$ \quad
  \textbf{Kate Knill}$^1$ \quad
  \textbf{Mark Gales}$^1$ \\[0.2cm]
  $^1$ALTA Institute, Department of Engineering, University of Cambridge \\
  $^2$Department of Computational Linguistics, University of Zurich \\[0.2cm]
  \tt{\{ayu20, mq227, mjfg, kmk1001\}@cam.ac.uk} \\
}

\begin{document}
\maketitle
\begin{abstract}
Spoken native language (L1) identification for second language learners has traditionally relied on acoustic attributes of speech. Comparatively little attention has been paid to analysing speech transcripts for L1 traces. In this paper, we present a study of L1 identification using both speech and corresponding transcripts from a spoken learner English corpus. We compare a WavLM-based audio L1 classifier with a zero-shot GPT-4o classifier operating on transcripts and analyse their predictions across proficiency levels using a CEFR-balanced evaluation set. Our experiments show that the audio classifier outperforms the transcript-based approach overall when full transcripts are available, although their relative strengths vary with learner proficiency. Audio performs better at A1, whereas text performs better from A2 to C1 with full transcripts. Transcript-based accuracy generally increases with proficiency, while acoustic performance remains comparatively stable. Restricting transcript length substantially reduces text-based accuracy, whereas audio is less sensitive to input length and performs relatively well with limited input. These findings provide insights into L1 identification from acoustic and textual representations of learner speech and how performance varies with proficiency and the amount of available language material.
\end{abstract}

\section{Introduction}

Native language identification (NLI) aims to infer a language
user's first language (L1) from their production in a second
language (L2). Much of the existing work has focused on
learner writing, using lexical choices, grammatical patterns,
and spelling errors as evidence of L1 background
\citep{malmasi2017report}. In spoken language, cross-linguistic
influence may also appear in pronunciation, rhythm, and
intonation \citep{piske2001factors}. Learner speech therefore
offers potentially complementary sources of evidence for NLI:
audio retains phonetic and prosodic information, while
transcripts provide more direct access to lexical and grammatical patterns.

The usefulness of these sources of evidence may vary with
learner proficiency. Some grammatical errors become less
frequent at higher proficiency levels, although developmental
patterns differ across error types
\citep{thewissen2013capturing}. L1-related pronunciation patterns may also change with
L2 learning and experience \citep{derwing2013development}. More proficient learners may additionally produce longer and more varied responses \citep{alzahrani2024utility}, providing different opportunities for identifying their L1. These changes make
it unclear whether L1 identification from audio and transcripts follows similar patterns across proficiency levels.

To investigate this question, we compare an audio L1
classifier based on WavLM \citep{chen2022wavlm} with
zero-shot large language models (LLMs) prompted for textual NLI \citep{uluslu2026impact}, operating on transcripts of the same learner English responses. We create BULATS-L1, a new dataset for spoken NLI with an evaluation set balanced jointly across six L1 backgrounds and five manually assigned CEFR levels, from A1 to C1. The transcripts were generated by an automatic speech recognition (ASR) model adapted to learner speech \citep{qian2025end}. This dataset allows us to examine overall identification performance and differences across L1 groups, and to compare proficiency-related patterns while holding the distribution of L1 backgrounds constant. Our results show that transcript-based NLI outperforms the audio classifier overall when full transcripts are available. However, this advantage depends on proficiency and transcript length. Audio performs better at A1, while text performs better from A2 to C1 with full transcripts. Restricting transcript length reduces overall text-based accuracy whereas audio is
less sensitive to input length and performs relatively well with limited input. 

\section{Related Work}

Shared tasks have established benchmarks for NLI from both written and spoken learner language. The 2017 NLI
Shared Task evaluated identification from essays, spoken
responses, and their combination
\citep{malmasi2017report}. Strong essay-based systems used support vector machines with character and word n-grams and syntactic features, often combined through ensembles or classifier stacking. In the speech track, the top-ranked systems combined transcript-based features with acoustic i-vectors. Transcript-only systems performed considerably worse, with the relatively short spoken responses identified as a possible contributing factor.
The report consequently highlighted the need to investigate NLI using longer transcripts.

More recent research has investigated pretrained language
models and prompted LLMs as alternatives to feature-based NLI systems \citep{ng2025leveraging,goswami2025multilingual}. Exposure to multilingual text \citep{kew2024turning} and examples of translation \citep{zhu2024multilingual} and grammatical error correction \citep{masciolini2025multigec} during pretraining may equip LLMs with knowledge of cross-linguistic correspondences and nonstandard language patterns that can subsequently support NLI. Zero-shot approaches then infer L1 through task instructions without training a dedicated classifier on labelled NLI examples \citep{uluslu2025investigating}. 

On the audio side, self-supervised learning has enabled transferable speech representations learned from unlabelled
recordings \citep{baevski2020wav2vec}. Their applicability to speaker identification and verification has been examined
through benchmarks such as SUPERB \citep{yang2021superb}, while WavLM demonstrates strong performance across speech and speaker-related tasks \citep{chen2022wavlm}. Building on these developments, we compare WavLM-based audio classification with zero-shot transcript-based NLI on the same learner responses, examining performance across proficiency levels and the effect of transcript length.

\section{Datasets}
\label{sec:datasets}

We use two learner English speech corpora with complementary roles. BULATS-L1 is the primary corpus for our L1 identification experiments, providing self-reported L1 labels
and a CEFR-balanced evaluation set, but no manual transcriptions.
The Speak \& Improve Corpus 2025 \citep{knill2024speak,knill25_slate} provides manual transcriptions for learner speech, allowing us to assess the sensitivity of transcript-based L1 predictions to automatic transcription.

\begin{table}[!htbp]
    \centering
    \begin{tabular}{lrr}
        \toprule
        CEFR level & $n$ & Avg. Length \\
        \midrule
        A1 & 54 & 190 $_{\pm88}$ \\
        A2 & 54 & 286 $_{\pm 88}$ \\
        B1 & 54 & 390 $_{\pm 94}$ \\
        B2 & 54 & 464 $_{\pm 80}$ \\
        C1 & 54 & 555 $_{\pm 101}$ \\
        \midrule
        Overall & 270 & 377 $_{\pm 157}$ \\
        \bottomrule
    \end{tabular}
    \caption{BULATS-L1 dataset composition by CEFR level.
    Each level includes nine submissions from each of six L1
    backgrounds. Transcript lengths are measured in words
    using the ASR generated transcripts. The values are reported as mean $\pm$ standard deviation.}
    \label{tab:dataset-composition}
\end{table}

\subsection{BULATS-L1}
\label{sec:BULATS-L1}

We construct the BULATS-L1 dataset from BULATS (Business Language Testing Service)\footnote{\url{https://en.wikipedia.org/wiki/Business_Language_Testing_Service}}, a discontinued online English language assessment service developed by Cambridge English. The dataset includes learners with six self-reported L1 backgrounds: Arabic, Chinese, French, Italian, Japanese, and Spanish. These L1 groups were selected based on the availability of sufficient test submissions across the included CEFR levels to construct a balanced evaluation set. 

\paragraph{Audio classifier training and development data.}
The BULATS-L1 training set comprises 16,149 utterances from 769 speakers across the six target L1 backgrounds, totalling approximately 86.4 hours of speech. The training and evaluation sets are speaker-disjoint. %Unlike the balanced evaluation set, the training set is imbalanced across L1 backgrounds, ranging from 32 Arabic-speaking learners (3.6 hours) to 213 Spanish-speaking learners (24.0 hours).
Unlike the balanced evaluation set, the training data could not be perfectly balanced across L1 backgrounds because of differences in data availability. We used as much available data as possible for the lower-resource L1s, particularly Arabic and Mandarin, while downsampling higher-resource L1s such as Spanish to reduce the disparity. As a result, the amount of training data ranges from 32 Arabic-speaking learners (3.6 hours) to 213 Spanish-speaking learners (24.0 hours). The BULATS-L1 development set has a similar distribution as the training set, with a total of 15.05 hours of speech across six target L1 backgrounds. The development set is used for hyperparameter turning and model selection only.

\paragraph{Balanced evaluation set.} Each BULATS test submission contains responses to five distinct speaking tasks, each targeting specific aspects of speaking proficiency. We exclude Part 2, a read-aloud task with prescribed linguistic content, and concatenate the remaining four spoken responses within each test submission for audio-based analysis and their corresponding transcripts for transcript-based analysis. The evaluation set is balanced jointly across L1 backgrounds and CEFR levels A1--C1, with nine test submissions per L1--CEFR combination, yielding 54 submissions per proficiency level and 270 submissions overall. C2 was excluded because insufficient submissions were available to meet this balancing criterion. CEFR proficiency levels are manually assigned by human experts.

Manual transcriptions are not available for this corpus. We generate transcripts using the Whisper-based ASR model adapted to learner speech by \citet{qian2025end}, and use these transcripts as input to the transcript-based L1 classifiers. Table~\ref{tab:dataset-composition} summarizes the evaluation set by CEFR level. Mean transcript length per submission increases from 190 words at A1 to 555 words at C1. Although submission counts are balanced, the amount of available text therefore varies across proficiency levels.

\subsection{Speak \& Improve 2025}
\label{sec:speak-improve}
The Speak \& Improve Corpus 2025 is a public dataset of second-language (L2) learner speech \citep{knill2024speak,knill25_slate}.
We use the fluent manual transcriptions from the evaluation subset of the Speak \& Improve Corpus 2025, comprising 300 test submissions. These transcripts retain learner language errors but exclude disfluencies. Since reference L1 labels are unavailable in the data used here, we assess L1 prediction consistency between manual and automatic transcriptions rather than L1 identification accuracy.

%updated
\begin{table*}[t]
    \centering
    \setlength{\tabcolsep}{12pt}
    \begin{tabular}{lcccc}
        \toprule
        & \multicolumn{2}{c}{Accuracy}
          & \multicolumn{2}{c}{Macro-F1} \\
        \cmidrule(lr){2-3}
        \cmidrule(lr){4-5}
        CEFR  & Audio & Text & Audio & Text \\
        \midrule
        A1  & 81.48 & 48.15 & 81.03  & 45.31 \\
        A2  & 83.33 & 87.04 & 81.31  & 87.06 \\
        B1  & 88.89 & 90.74 & 88.45  & 90.58 \\
        B2  & 92.59 & 94.44 & 92.46  & 94.30 \\
        C1  & 81.48 & 92.59 & 81.25  & 92.67 \\
        \midrule
        Overall & 85.56 & 82.59 & 85.37 & 82.60 \\
        \bottomrule
    \end{tabular}
    \caption{Submission-level NLI accuracy and macro-F1
    (\%) on BULATS-L1 by CEFR level. Text results are unrestricted transcripts.}
    \label{tab:submission-nli-cefr-attention-audio}
\end{table*}

\section{Experimental Setup}

\subsection{Speech-based NLI}
\label{sec:approach_audio_l1}

For speech-based NLI, we train a classifier on representations from a frozen pretrained WavLM-Large~\footnote{https://huggingface.co/microsoft/wavlm-large} encoder \citep{chen2022wavlm}. Audio is sampled at 16\,kHz. We learn a weighted combination of the encoder's hidden representations rather than using only the final transformer layer:
\begin{equation}
\mathbf{H} = \sum_{l=0}^{L} \alpha_l \mathbf{H}^{(l)},
\end{equation}
where $\mathbf{H}^{(l)}$ denotes the representation at layer $l$, $L$ is the number of transformer layers, and $\alpha_l$ are softmax-normalised trainable layer weights.

We apply attention pooling over valid encoder frames to obtain a clip-level embedding. The classification head consists of a fully connected layer with ReLU activation and dropout, followed by a linear output layer. A softmax converts the output logits into probabilities over the six L1 classes. Only the layer weights and classification head are trained, using AdamW with cross-entropy loss and label smoothing. Training hyperparameters are provided in Appendix~\ref{app:audio-hyperparameters}.

At inference time, we divide each concatenated test submission into consecutive, non-overlapping 30-second clips, retaining any shorter final clip. We process all clips and average their predicted class probabilities to obtain a submission-level distribution:
\begin{equation}
p(y \mid x) =
\frac{1}{K} \sum_{k=1}^{K} p(y \mid x_k),
\end{equation}
where $K$ is the number of clips in submission $x$. The class with the highest average probability is selected as the predicted L1.

\subsection{Transcript-based NLI}
\label{sec:approach_text_l1}

For transcript-based NLI, we follow the zero-shot classification procedure of \citet{uluslu2026impact}, using OpenAI's \texttt{gpt-4o}. We formulate NLI as a multiple-choice task, mapping
the six candidate L1 labels to letter identifiers. The prompt templates are provided in Appendix~\ref{sec:prompts}. Constrained decoding restricts the response to valid identifiers, and their
token-level log probabilities are normalised to obtain a distribution over the candidate classes.

To reduce sensitivity to label position, we evaluate each transcript using three randomised candidate orderings \citep{liusie2024llm}. We map the resulting
probabilities back to their corresponding L1 labels and average the distributions. The highest-probability class is selected as the prediction. Before classification, we mask named entities, including
locations, nationalities, and personal names, as well as non-English words. This approach aims to reduce reliance on explicit identity cues and topic-related shortcuts \citep{uluslu2025robust}.

\subsection{Evaluation}
\label{sec:evaluation}

\paragraph{Sensitivity to automatic transcription.}
On Speak \& Improve, we compare fluent manual transcripts and ASR-generated transcripts. We calculate WER separately for each test submission and report the mean across submissions. To assess the effect on transcript-based NLI, we report predicted-label agreement, defined as the proportion of paired transcripts receiving the same predicted L1, and mean Jensen--Shannon divergence (JSD) \citep{lin1991divergence} between their predicted L1 distributions. Agreement captures changes in the selected label, while JSD captures differences across the full distribution. Since reference L1 labels are unavailable, these metrics assess prediction consistency rather than L1 identification accuracy.

\paragraph{L1 identification.}
On BULATS-L1, we evaluate both pipelines against self-reported L1 labels at the test-submission level. We report accuracy and macro-averaged F1. Macro-averaged metrics assign equal weight to each of the six L1 classes. Macro-F1 is computed by averaging the class-specific F1-scores. %We additionally report performance by L1 background and CEFR level to examine differences between the two pipelines.

\begin{figure*}[!t]
    \centering
    \includegraphics[width=0.95\linewidth]{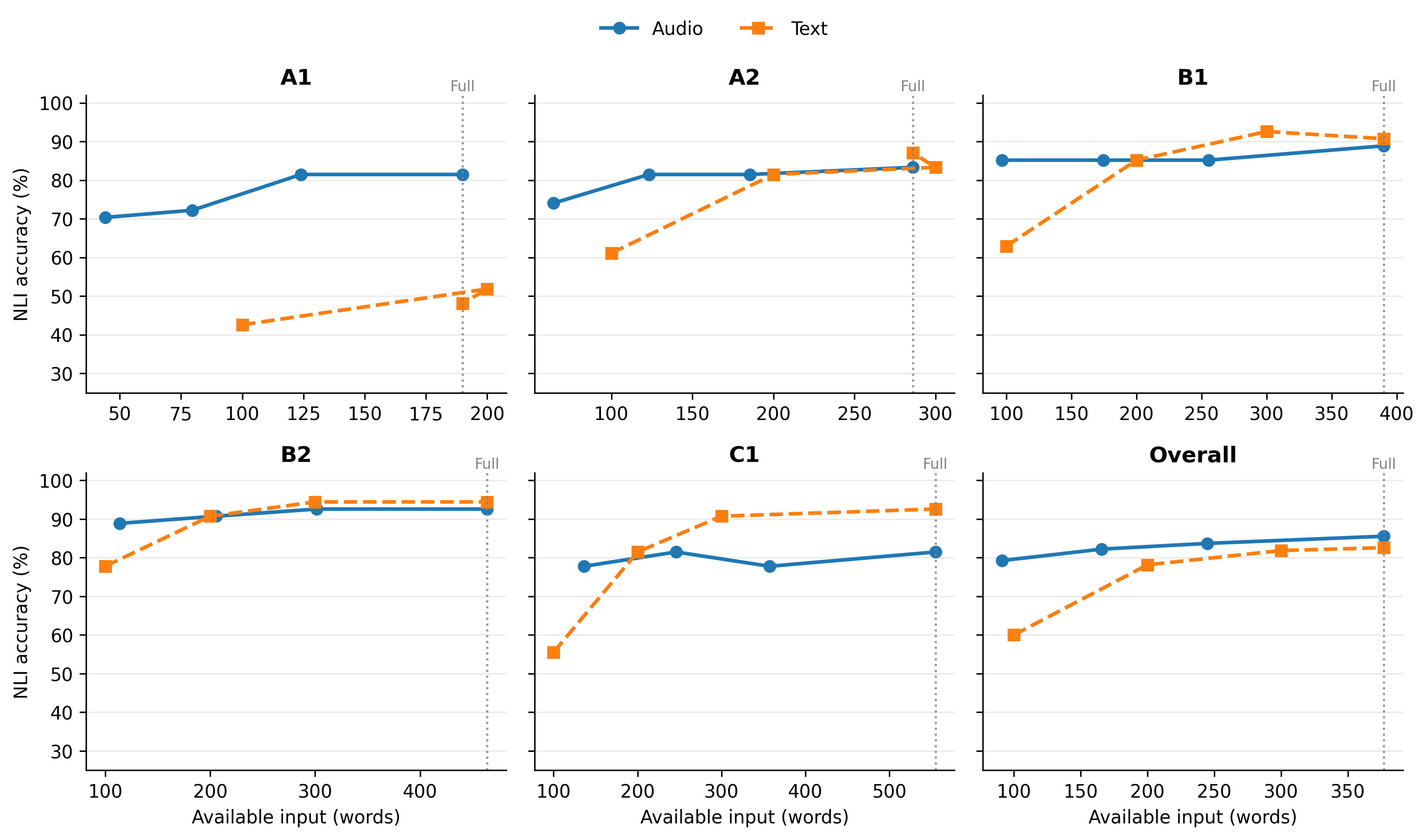}
    \caption{Effect of available input on audio- and text-based NLI}
    \label{fig:nli_input_length}
\end{figure*}

\section{Results}
\subsection{NLI across Proficiency Levels}
Table~\ref{tab:submission-nli-cefr-attention-audio} compares the speech-based and transcript-based systems on BULATS-L1. Using unrestricted transcripts, GPT-4o achieves an overall accuracy of 82.59\%, compared with 85.56\% for the WavLM-based classifier, a difference of 2.97 percentage points. The corresponding macro-F1 scores are 82.60\% and 85.37\%, respectively, showing a similar overall pattern. % Updated audio performance

The relative performance of the two systems varies across proficiency levels. At A1, the audio classifier clearly outperforms the transcript-based system, achieving 81.48\% accuracy compared with 48.15\%, a different of 33.33 percentage points. From A2 onwards, however, the transcript-based system achieves higher accuracy. Its accuracy rises sharply from 48.15\% at A1 to 94.44\% at B2, before decreasing slightly to 92.59\% at C1. Audio accuracy varies over a narrower range, increasing from 81.48\% at A1 to 92.59\% at B2 before dropping to 81.48\% at C1.
Macro-F1 closely follows the same trend as accuracy across proficiency levels, suggesting that the observed differences are not driven by a small subset of L1 classes.

These results show different proficiency-related patterns for the two pipelines. Transcript-based identification improves substantially between A1 and B2 by 46.29\% accuracy, whereas audio performance is comparatively stable. %Although audio accuracy decreases from B2 to C1, this result alone does not establish that accent-related information has diminished. 
Both systems decline at C1, although the decrease is considerably larger for the audio classifier. This pattern alone does not establish that L1-related acoustic information diminishes at higher proficiency, since other factors may also contribute.
In particular, transcript length increases with proficiency, motivating a closer examination of the effect of available text on identification performance.

\subsection{Effect of Transcript Length}
\label{sec:results-length}
To examine the contribution of available text to
transcript-based NLI, we evaluate transcripts capped at 100, 200, and 300 words alongside the unrestricted condition. For each capped condition, we retain the first $N$ words of each concatenated transcript, leaving
shorter transcripts unchanged. Table~\ref{tab:text-length-results} reports
accuracy across these conditions.

%updated
\begin{table}[!htbp]
    \small
    \centering
    \begin{tabular}{l|ccccc|c}
    \toprule
    Len. & A1 & A2 & B1 & B2 & C1 & All \\
    \midrule
        100 & 42.59 & 61.11 & 62.96 & 77.78 & 55.56 & 60.00 \\
        200 & 51.85 & 81.48 & 85.19 & 90.74 & 81.48 & 78.15 \\
        300 & 48.15 & 83.33 & 92.59 & 94.44 & 90.74 & 81.85 \\
        Full & 48.15 & 87.04 & 90.74 & 94.44 & 92.59 & 82.59 \\
    \bottomrule
    \end{tabular}
    \caption{Transcript-based NLI accuracy (\%) by CEFR
    level and maximum transcript length in words.}
    \label{tab:text-length-results}
\end{table}

Overall accuracy increases with the word cap, from 60.00\% at 100 words to 78.15\% at 200 words and 81.85\% at 300 words, with a further small increase to 82.59\% for the full transcript. %Restricting the available text therefore substantially reduces identification performance overall, although improvements are not uniform within every proficiency level. For example, B2 accuracy decreases between the 100- and 200-word conditions before increasing at 300 words.
The largest gain occurs between 100- and 200-word conditions, while performance starts to saturate beyond 200 words.

Proficiency-related differences remain under a common word cap. With a 200-word cap, accuracy is 51.85\% at A1 and 81.48\% at A2, compared with 85.19\% at B1, 90.74\% at B2, and 81.48\% at C1. % Higher-proficiency groups therefore remain more accurately classified than the A1-A2 group even when the maximum amount of text is restricted. 
Thus, the substantially lower performance at A1 cannot be explained solely by the shorter responses typically produced by lower-proficiency learners. From A2 onwards, the classifier achieves relatively strong performance once sufficient text is available, suggesting that the availability or distinctiveness of lexical and grammatical L1 cues may also vary with proficiency.

% These findings suggest that the choice between the two pipelines depends partly on the amount of available evidence. The audio system outperforms all capped-text conditions in overall accuracy, whereas unrestricted text achieves the highest overall accuracy. However, the audio system uses complete recordings in all comparisons. Pronunciation patterns may recur within relatively short speech samples, while informative lexical and grammatical patterns may require more extensive language production, but our experiments do not directly test this distinction. Evaluating sensitivity to audio duration and comparing the two pipelines on matched speech segments remain directions for future work.
The strong dependence of transcript-based NLI on input length motivates a corresponding analysis of the audio system. Although the full audio system achieves higher overall accuracy, the two modalities may differ in how performance depends on input length. %We therefore examine the effect of restricting the audio input and compare it with the text results.

\begin{table}[!b]
    \centering
    \small
    \begin{tabular}{@{ }l@{ }|p{5.mm}|p{4.8mm}p{5mm}p{5mm}p{5mm}p{5.8mm}|p{4.8mm}}
    \toprule
    Parts & Len & A1 & A2 & B1 & B2 & C1 & All \\
    \midrule
    P1      & 91  & 70.37 & 74.07 & 85.19 & 88.89 & 77.78 & 79.26 \\
    P1,3    & 166 & 72.22 & 81.48 & 85.19 & 90.74 & 81.48 & 82.22 \\
    P1,3,4  & 245 & 81.48 & 81.48 & 85.19 & 92.59 & 77.78 & 83.70 \\
    P1,3,4,5 &377 & 81.48 & 83.33 & 88.89 & 92.59 & 81.48 & 85.56\\
    \bottomrule
    \end{tabular}
    \caption{Audio NLI accuracy (\%) on restricted audio parts by CEFR level and average length in words.}
    \label{tab:audio_length}
\end{table}

\subsection{Effect of Audio Length}
\label{sec:audio_length}
We examine the effect of available audio by progressively adding the retrained task responses: P1, P1 + P3, P1 + P3 + P4, and P1 + P3 + P4 + P5. Table~\ref{tab:audio_length} reports accuracy across these conditions. Figure~\ref{fig:nli_input_length} plots audio- and text-based model accuracy overall and at each CEFR as the amount of available input words increases. Per-part performance is in Appendix~\ref{app:audio-per-part}.

Audio-based NLI is less sensitive to input length than transcript-based NLI in overall performance. Audio accuracy increases from 79.26\% using only P1 to 82.22\% after adding P3, and reaches 85.56\% with all four retained parts. In comparison, text accuracy increases from 60.00\% at 100 words to 78.15\% at 200 words and 82.59\% with the full transcript. The audio system therefore achieves relatively strong performance with limited input, while the text system benefits more substantially from additional text material.

The same general pattern is observed across proficiency levels, although the effect varies by CEFR level. At A1, audio consistently outperforms text across the available input range. At A2, the initial advantage of audio narrows as more text becomes available, and the text system slightly exceeds audio with the full input. A similar crossover occurs for B1 and B2, where the text model outperforms the audio model once sufficient transcript material is available. At C1, text accuracy increases substantially with additional input and substantially outperforms the audio model, whereas the audio model performance varies within a comparatively narrow range.

These results suggest different sensitivities to input length across the two modalities. Acoustic L1 cues appear to be informative from relatively short speech samples, with additional audio providing modest gains. In contrast, transcript-based NLI benefits more from having longer transcripts, especially at shorter input lengths. The remaining differences across CEFR levels indicate that input length alone does not explain the proficiency-related differences; the availability and strength of acoustic and lexical L1 cues may also vary with proficiency.

\subsection{Fusion of Audio and Text Predictions}
\label{sec:audio_length}

\begin{figure}[!t]
    \centering
    \includegraphics[trim=0 5mm 0 0mm, width=0.9\linewidth]{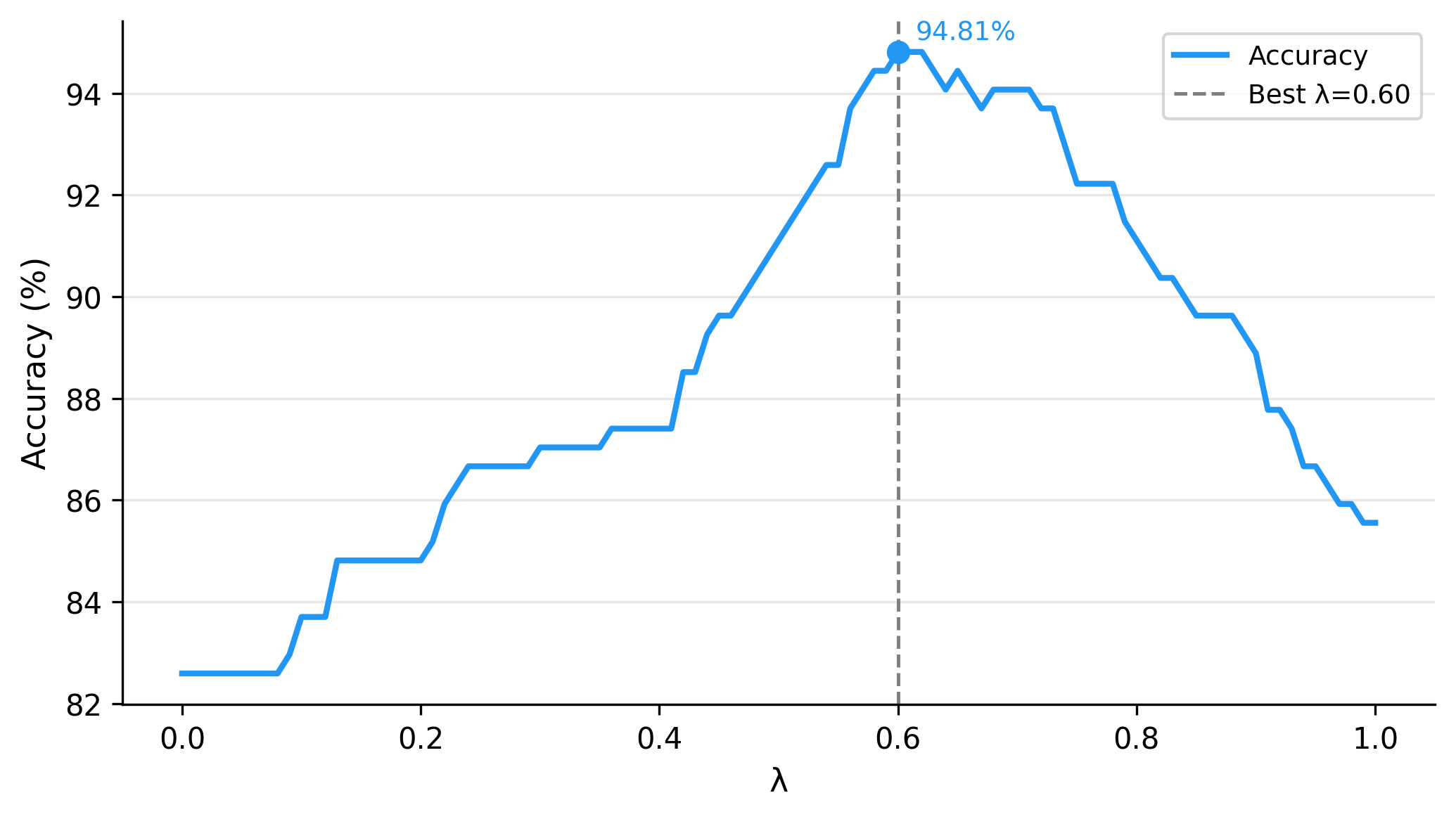}
    \caption{NLI accuracy on BULATS-L1 for linear fusion of audio and text predictive distributions}
    \label{fig:fusion}
\end{figure}

We further examine whether the audio and text modalities provide complementary information by linearly interpolating the predicted L1 probability distributions at the submission-level,
\begin{equation}
    p_{\mathrm{fuse}} = \lambda p_{\mathrm{audio}}
    + (1-\lambda)p_{\mathrm{text}},
\end{equation}
where $\lambda$ controls the weight of the audio model. Figure~\ref{fig:fusion} shows accuracy as $\lambda$ varies from 0 to 1.

Fusion consistently improves over either individual system for a broad range of weights. The best performance is obtained at $\lambda=0.60$, reaching 94.81\% accuracy, compared with 85.56\% for audio alone and 82.59\% for text alone. Performance peaks when slightly more weight is assigned to the audio model, but remains high around the optimum. These results indicate that the audio- and transcript-based systems capture complementary L1 information, and that combining their predictions substantially improves identification performance.

\subsection{Prediction Consistency on Speak \& Improve}

% On Speak \& Improve, the automatic transcripts yield 
We use Speak \& Improve to examine prediction consistency in the absence of reference L1 labels. We first assess the sensitivity of transcript-based NLI to automatic transcription. The ASR transcripts have
a mean submission-level WER of 9.2\% against fluent manual references. The transcript-based NLI classifier produces the same predicted L1 under both the manual and automatic transcription conditions for 89.7\% of submissions, with a mean JSD of 0.075 between the predictive distributions. Thus, the predicted label remains unchanged for most submissions, although approximately one in ten predictions differs. The high agreement suggests that automatic transcription retains information used by the classifier for L1 prediction in most cases, although it does not establish the accuracy of those predictions.

% agreement results
We then compare the audio- and transcript-based classifiers.
The two models produced the same top-ranked L1 for 77.7\% of the 300 submissions, with a mean Jensen--Shannon divergence (JSD) of 0.221 bits between their six-class predictive distributions. This indicates substantial label agreement alongside moderate differences in prediction probabilities.

\section{Conclusion}
\label{sec:conclusion}

We compared a WavLM-based audio classifier with a zero-shot GPT-4o transcript-based classifier for NLI using BULATS-L1, whose evaluation set is balanced across six L1 backgrounds and five manually assigned CEFR levels. %With full transcripts, the text system achieved higher overall accuracy than the audio system. Their relative performance varied with proficiency: audio performed better at A1, while text performed better from A2 to C1. Restricting transcript length substantially reduced overall text-based accuracy.
The audio system achieves higher overall accuracy than the transcript-based system, while their relative performance varies with proficiency. Audio performs substantially better at A1, whereas text is stronger at the higher proficiency levels once sufficient transcript material is available. Both modalities are affected by input length, although the audio system is less sensitive to shorter input than the transcript-based system.

% Overall, the results indicate that both learner proficiency and the amount of speech material matter when comparing audio-based and transcript-based NLI. Future work will examine audio-duration effects, compare predictions on matched speech segments, and investigate whether combining the two pipelines improves identification across proficiency levels.
The results also show that the two modalities provide complementary information. Linear fusion of their predictive distributions improves performance substantially, reaching 94.81\% accuracy, compared with 85.56\% for audio alone and 82.59\% for text alone. Overall, the findings indicate that learner proficiency, input length, and modality all influence NLI performance, and that combining acoustic and lexical evidence provides the strongest identification results.

\section*{Acknowledgments}
This work reports on research supported by Cambridge University Press \& Assessment, a department of The Chancellor, Masters, and Scholars of the University of Cambridge. It was also supported by Innosuisse innovation project 103.188 IP-ICT (AUCH).

\bibliographystyle{acl_natbib}
\bibliography{nlp4call}

\newpage
\appendix

\section{L1 Classification Prompt}
\label{sec:prompts}
For the L1 attribution task, the model was cast in an expert persona and provided with a multiple-choice selection of the six candidate languages (represented by \textit{\{options\}}). The exact prompt template was:
\begin{quote}
``You are a forensic linguistics expert that reads essays from non-native speakers of English in order to identify their native language. Use clues such as word choice, syntactic patterns, and grammatical errors to decide. Analyze the text and choose the most likely native language from the following options:

\{options\}

(X) Other

Reply with only the single letter (A, B, C, etc.) representing your choice.''
\end{quote}

\section{Audio Model Parameters}
\label{app:audio-hyperparameters}
The audio classifier was trained for 10 epochs with the WavLM-Large encoder frozen throughout training; only the layer-combination weights and classification head were updated. We used AdamW with a learning rate of $1\times10^{-4}$ and weight decay of $10^{-2}$. The learning rate followed a cosine schedule with 500 warm-up steps. Cross-entropy loss with label smoothing of 0.1 was used as the training objective. The batch size was 2, with gradients accumulated over four steps, resulting in an effective batch size of 8. Training used mixed-precision computation, and gradients were clipped to a maximum norm of 1.0. The random seed was set to 42. Model performance was evaluated on the development set after each epoch, and the checkpoint achieving the highest development-set accuracy was retained for evaluation.

\section{Audio Model Per-part Performance}
\label{app:audio-per-part}
\begin{table}[!htbp]
    \centering
    \small
    \begin{tabular}{l|ccccc|c}
    \toprule
    Part & A1 & A2 & B1 & B2 & C1 & All \\
    \midrule
    P1 & 70.37 & 74.07 & 85.19 & 88.89 & 77.78 & 79.26 \\
    % P2 & & & 85.19 & 87.04 & 83.33 & 83.33 & 66.67 & 81.11  \\
    P3 & 75.93 & 79.63 & 87.04 & 92.59 & 87.04 & 84.44 \\
    P4 & 79.63 & 85.19 & 92.59 & 87.04 & 79.63 & 84.81 \\
    P5  & 74.07 & 79.63 & 92.59 & 90.74 & 79.63 & 83.33 \\
  \bottomrule
    \end{tabular}
    \caption{Audio NLI accuracy on per-part audios}
\end{table}

\end{document}